\documentclass[a4paper]{cas-sc}

\usepackage[numbers]{natbib}

\usepackage{subfigure}
\usepackage{amsmath,amssymb,amsfonts}
\usepackage{algorithmic}
\usepackage{graphicx}
\usepackage{textcomp}
\usepackage{xcolor}
\usepackage{multirow}

\begin{document}
\let\WriteBookmarks\relax
\def\floatpagepagefraction{1}
\def\textpagefraction{.001}

\shorttitle{Unsupervised Network for Single Image Raindrop Removal}    

\shortauthors{H. Wang et al.}  

\title [mode = title] {Unsupervised Network for Single Image Raindrop Removal}  

\tnotemark[1] 

\tnotetext[1]{The research was partially supported by the National Natural Science Foundation of China under Grants 62271354 and 61871297.} 

%

\author[1]{Huijiao Wang}[orcid=0000-0002-8113-091X]



\ead{huijiao_wang@whu.edu.cn}


\credit{Conceptualization of this study, methodology, coding, doing experiments, and part of the writing, including abstract, experiments,  and conclusion }

\author[2]{Shenghao Zhao}[]


\ead{e0724582@u.nus.edu.sg}


\credit{Doing part of the experiments and part of the writing }



\author[1]{Lei Yu}[]

\ead{ly.wd@whu.edu.cn}


\credit{Conceptualization of this study, paper editing}

\cormark[1]
\cortext[cor1]{Corresponding author}


\author[3]{Xulei Yang}[]

\ead{yang_xulei@i2r.a-star.edu.sg}


\credit{Conceptualization of this study, supervision, and writing}


\affiliation[1]{organization={Wuhan University},
            addressline={Luojia Hill, Wuchang}, 
            city={Wuhan},
            postcode={430072}, 
            state={Hubei},
            country={China}}

\affiliation[2]{organization={National University of Singapore (NUS)},
            country={Singapore}}

\affiliation[3]{organization={Institute for Infocomm Research (I$^2$R), A*STAR},
            country={Singapore}}


\begin{abstract}
Image quality degradation caused by raindrops is one of the most important but challenging problems that reduce the performance of vision systems. Most existing raindrop removal algorithms are based on a supervised learning method using pairwise images, which are hard to obtain in real-world applications. This study proposes a deep neural network for raindrop removal based on unsupervised learning, which only requires two unpaired image sets with and without raindrops. Our proposed model performs layer separation based on cycle network architecture, which aims to separate a rainy image into a raindrop layer, a transparency mask, and a clean background layer. The clean background layer is the target raindrop removal result, while the transparency mask indicates the spatial locations of the raindrops. In addition, the proposed model applies a feedback mechanism to benefit layer separation by refining low-level representation with high-level information. i.e., the output of the previous iteration is used as input for the next iteration, together with the input image with raindrops. As a result, raindrops could be gradually removed through this feedback manner. Extensive experiments on raindrop benchmark datasets demonstrate the effectiveness of the proposed method on quantitative metrics and visual quality. 
\end{abstract}




\maketitle











\section{Introduction}
\label{sec:intro}
Images captured in rainy weather face significant degradation due to raindrops, presenting a critical challenge in various applications, especially those involving driving scenes such as driver assistance systems and autonomous driving. Raindrops adhering to the lens, windscreen, or glass windows not only blur parts of the background but also profoundly influence overall image quality. Consequently, there is a pressing need for an effective method capable of automatically removing raindrops and restoring the pristine background.

This paper focuses on the intricate task of recovering a clean background from a single image plagued by raindrops. Single-image raindrop removal proves challenging due to the inherent diversity between images with and without raindrops. Detecting raindrops accurately is difficult owing to variations in their shapes, sizes, and densities. Moreover, raindrops introduce background distortion, complicating the retrieval of obscured information. The presence of raindrops on the lens further exacerbates the problem by altering the focus of both foreground and background, leading to unexpected blurring.

\begin{figure}[!htbp]
\centering
\subfigure[]{
\begin{minipage}[t]{0.2\linewidth}
\centering
\includegraphics[height=0.85in]{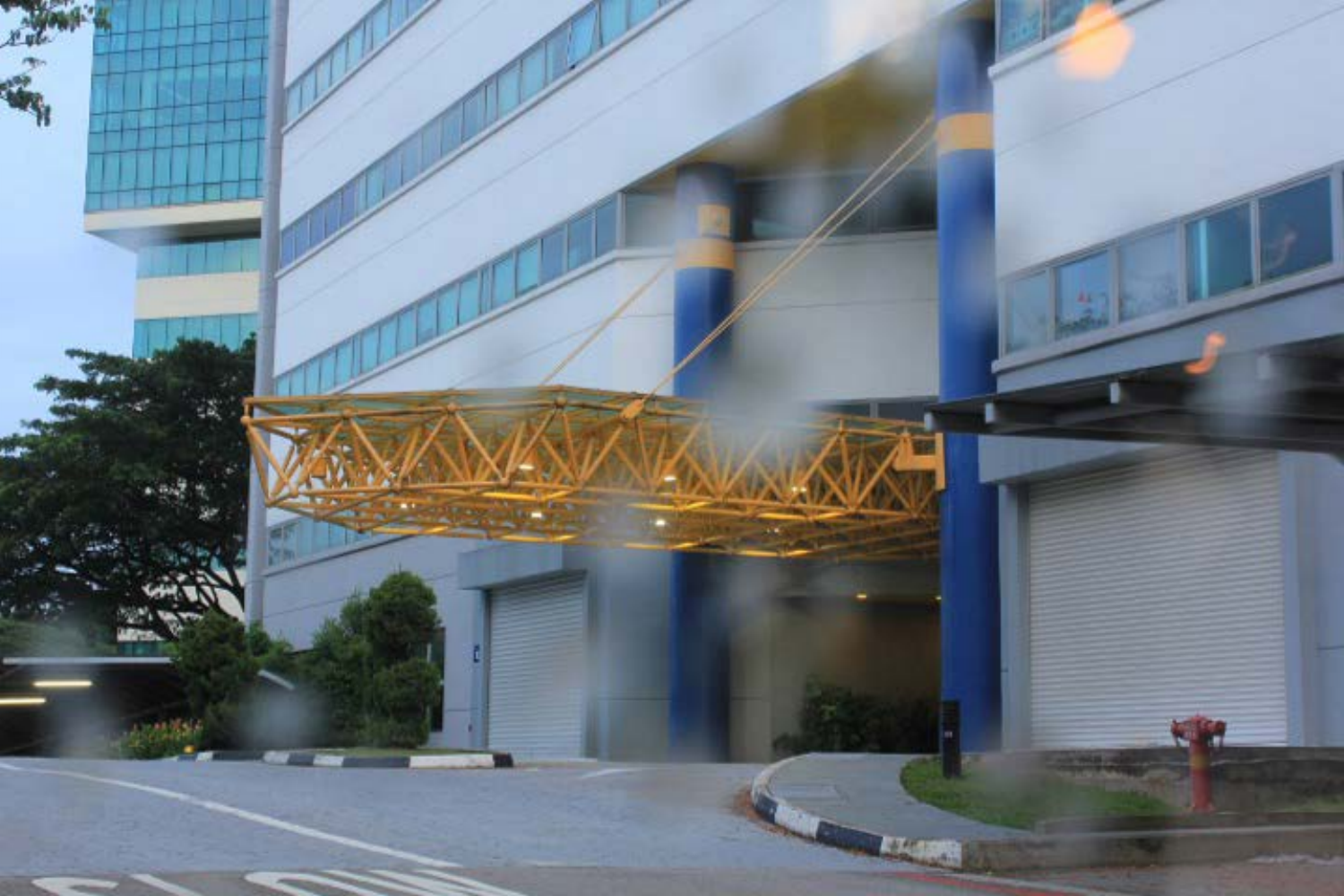}
\end{minipage}%
}%
\subfigure[]{
\begin{minipage}[t]{0.2\linewidth}
\centering
\includegraphics[height=0.85in]{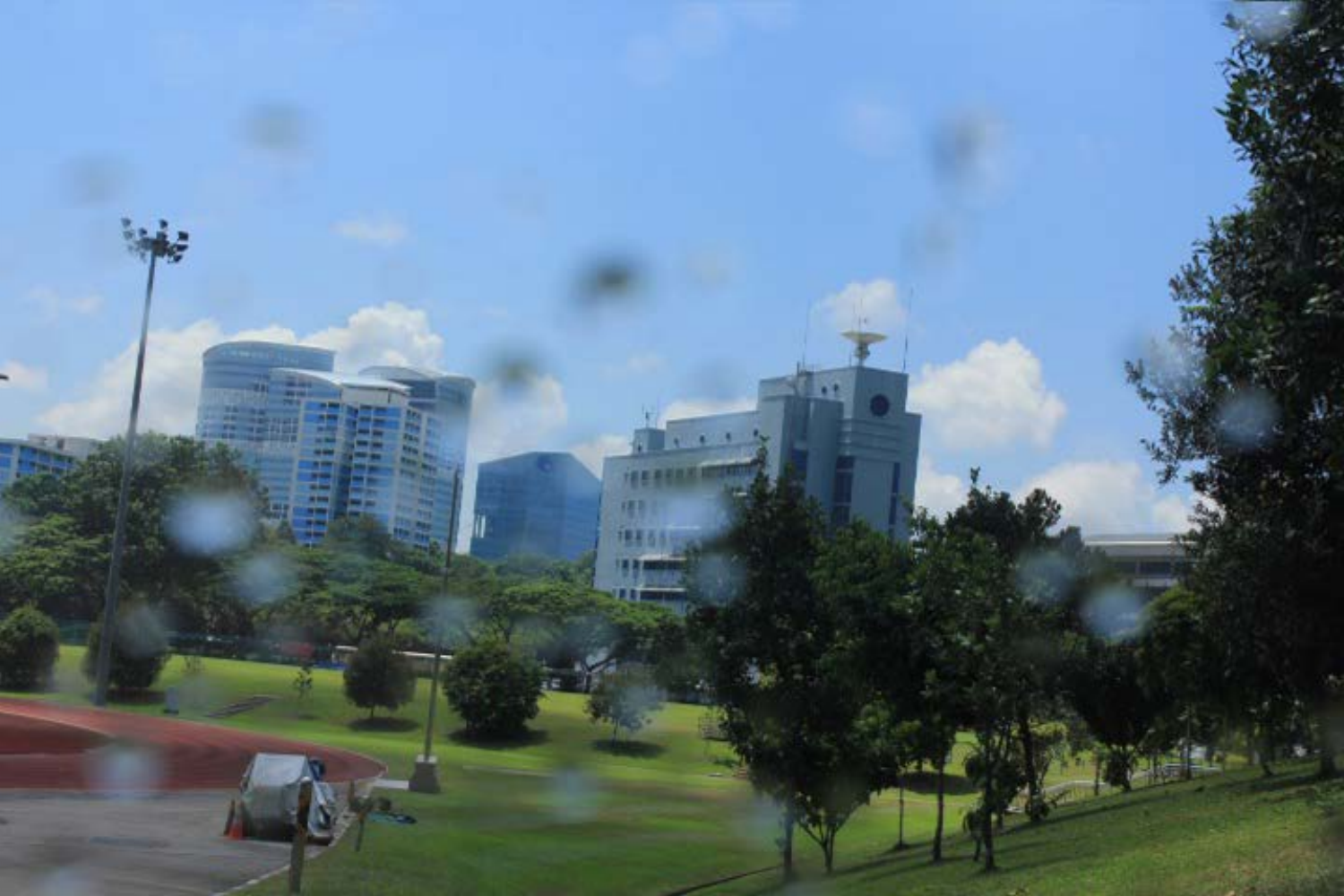}
\end{minipage}%
}%
\subfigure[]{
\begin{minipage}[t]{0.23\linewidth}
\centering
\includegraphics[height=0.85in]{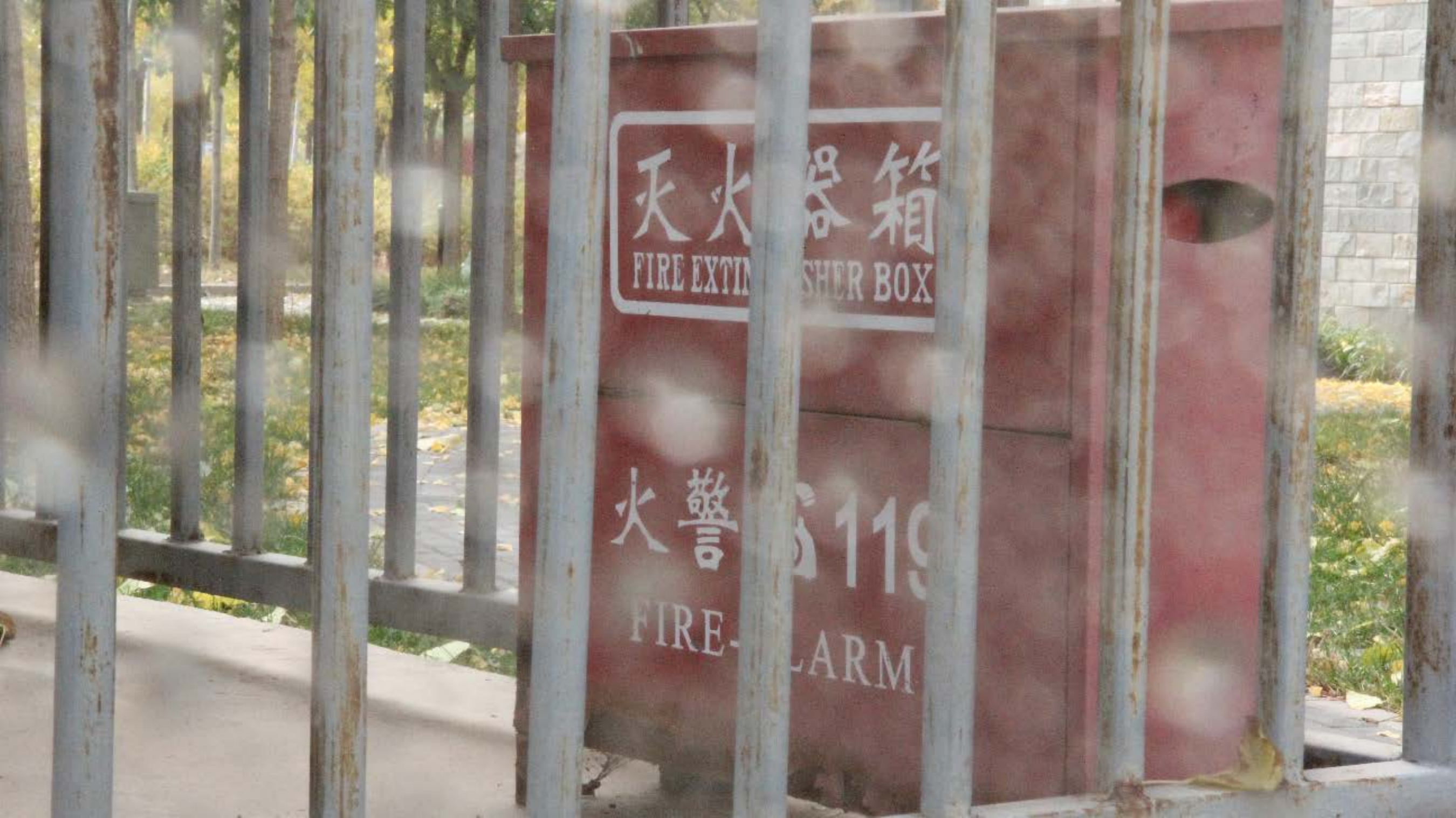}
\end{minipage}
}%
\subfigure[]{
\begin{minipage}[t]{0.23\linewidth}
\centering
\includegraphics[height=0.85in]{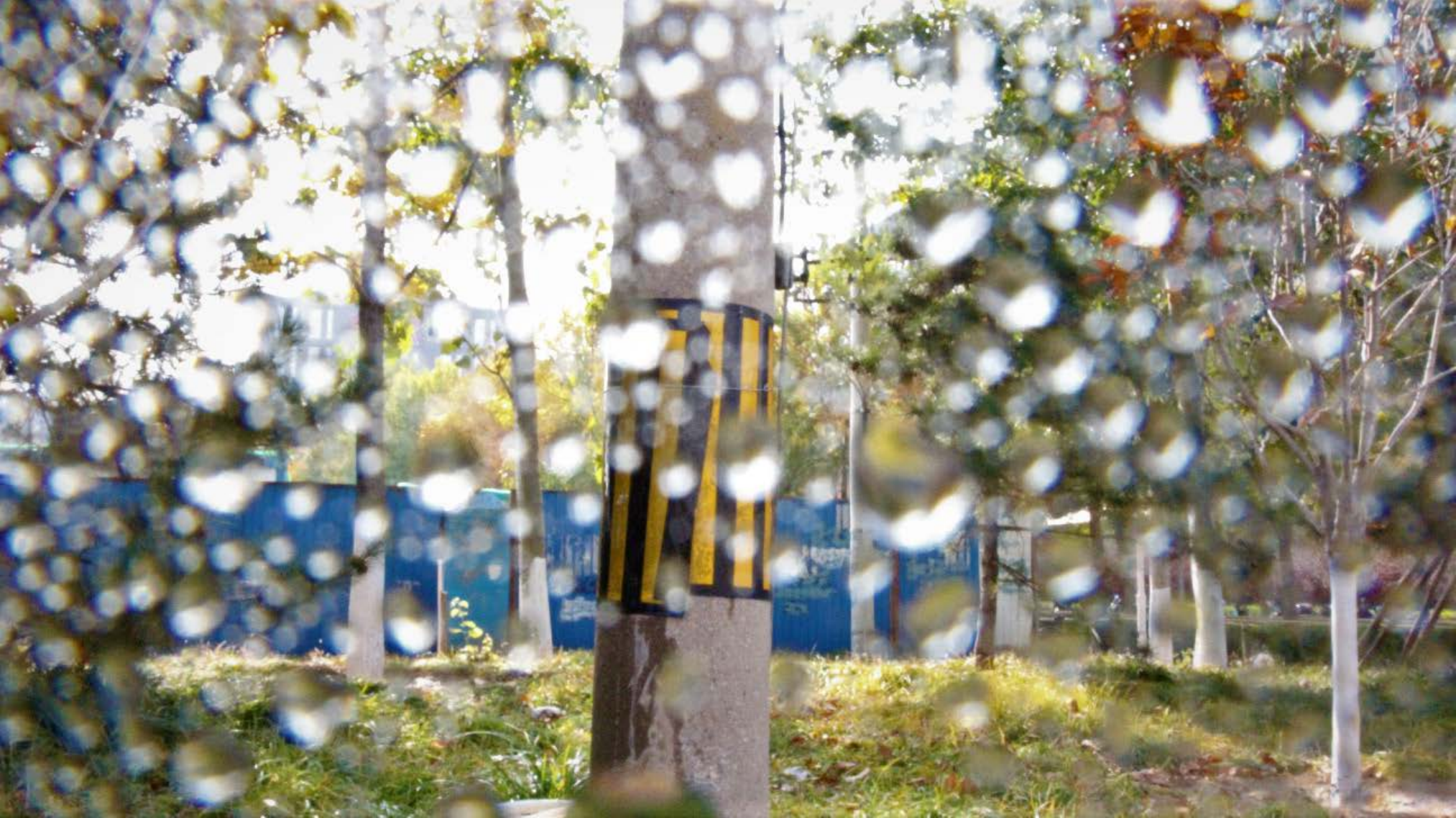}
\end{minipage}
}%
\centering
\caption{Examples of rainy images in the datasets we used. (a) and (b) are from the NUS raindrop dataset \cite{qian2018attentive}. 
(c) and (d) are from RainDS \cite{Quan2021InOneGo}.}
\label{fig1}
\end{figure}

Traditionally, researchers have proposed model-based methods \cite{roser2009video,roser2010realistic, xu2012improved,kim2013single} to address raindrop removal. However, recent attention has shifted towards deep-learning-based models, which have shown greater promise than their traditional counterparts. Numerous raindrop removal methods \cite{qian2018attentive, Eigen2013dirt, Li2020AllInOne, Quan2021InOneGo, Hao2019Synthetic, Quan2019}, leveraging supervised learning, have demonstrated significant progress. Notably, Qian \emph{et al.} \cite{qian2018attentive} and RainDS dataset \cite{Quan2021InOneGo}  have shared paired datasets for raindrop simulation. It contains synthetic images based on public autonomous driving datasets. Fig. \ref{fig1} shows some examples of rainy images from the above datasets. Despite their utility, these datasets have limitations in fully simulating real-world raindrop characteristics, including shape, density, and size. Yan \emph{et al.} \cite{Yan2022RainGAN}  propose an unsupervised method to address this challenge. They posit that a raindrop comprises a clean image and a raindrop style, introducing a domain-invariant residual block to learn a raindrop-style latent code. Through the composition of a clean content image and various raindrop-style codes for data augmentation, their method aims to recover a clean image from a raindrop image. However, simply learning the shape of raindrops from a limited data set without considering its structural characteristics at different levels is not enough to deal with the raindrop removal problem in the real world.

In response to the challenges posed by real-world single-image raindrop removal, this paper presents a pioneering unsupervised image decomposition method coupled with a feedback network. Diverging from conventional supervised learning paradigms, our innovative approach relies solely on unpaired datasets – one comprising pristine images and the other featuring raindrop-laden counterparts. Taking inspiration from the principles elucidated in Double-DIP \cite{Gandelsman2019DoubleDIP}, we integrate a layer separation concept to deconstruct a rainy image into distinct components: a clean background, a transparency mask, and a raindrop image. To systematically capture the structural intricacies of raindrops across varying scales within a single input, we introduce a feedback strategy. This strategic approach refines low-level information with high-level details in a top-down fashion, facilitating the gradual acquisition of raindrop knowledge from coarse to fine granularity. The feedback mechanism, central to this progressive learning process, ensures a comprehensive and nuanced understanding of raindrops. Furthermore, our model capitalizes on the robust unsupervised framework provided by CycleGAN \cite{Zhu2017Unpaired}. This architecture empowers our model with the capacity for unsupervised training, enhancing its adaptability and performance.

The primary contributions of this work can be summarized as follows:

\begin{itemize}
\item Introduction of a novel unsupervised image decomposition network for effectively separating a rainy image into a clean background, a transparency mask, and a raindrop image without requiring paired data.
\item Utilization of a feedback mechanism through iterative neural networks to progressively learn raindrop representations from coarse to fine, culminating in the acquisition of the desired droplet layer.
\item Conducting extensive experiments on benchmark raindrop datasets, resulting in competitive performance among unsupervised methods.
\end{itemize}

\section{Related works}
\label{sec:related}
Reconstructing the clean background image from a rainy image has been studied extensively for decades. Recovering clean images from rainy video sequences has been widely explored \cite{garg2005camera, bossu2011rain, barnum2010analysis, brewer2008shape, zhang2006rain, garg2007vision, garg2006photorealistic, garg2004detection}. Recently most works \cite{Valanarasu_2022_CVPR, Unsupervised_2023_TPAMI, Ye_2022_CVPR, Yu_2022_CVPR, Yu_2023_TMM, Single_2020_TPAMI, Yang_2020_TPAMI} focus on rain streak removal. Due to the variable shape of adherent raindrops and the distortion to the background, very few raindrop removal methods have been proposed. From the input perspective, the raindrop removal problem can be divided into two categories: single-image and multi-image. Most existing methods for this problem are based on multi-image methods, which aim at the joint removal of raindrops in multiple images from stereo or video through the strong correlation between multiple images. Roser and Geiger \cite{roser2009video} tried to detect raindrops with a photometric model, which improves accuracy utilizing adjacent image frames. You \emph{et al.} \cite{you2015adherent} took advantage of the information difference of pixels on the time scale to detect and remove raindrops through video completion techniques. Yan \emph{et al.} \cite{Yan2022Feature} employs a single image module and a multiple frame module to remove raindrops and recover the background using a series of unsupervised losses and self-learning from video data without ground truths. However, they also rely on the temporal consistency between adjacent frames. Due to the lack of temporal and stereo-optical information, raindrop removal from a single image in an unsupervised manner is still challenging.

Many researchers have studied the problem of raindrop removal with single-image methods based on deep-learning technology in recent years. 
They have made substantial progress, including supervised learning methods and unsupervised methods. We will review the most related works in the following.

\subsection{Supervised learning raindrop removal}
Since Eigen \emph{et al.} \cite{Eigen2013dirt} first proposed a deep neural network for raindrop removal, raindrop removal from a single image has attracted the attention of many researchers. There are some works \cite{qian2018attentive, Li2020AllInOne, Quan2021InOneGo, Hao2019Synthetic, Quan2019, Adherent_2022_Neurocomputing, Raindrop_2022_SIVP, Laplacian_2022_PR} investigating supervised learning methods. For example, Qian \emph{et al.} \cite{qian2018attentive} made a pairwise data set and proposed a method combining generative adversarial networks with attention modules, while Hao \emph{et al.} \cite{Hao2019Synthetic} introduced a detection network to learn the raindrop refraction and blurring with paired synthetic photorealistic data. More recently, Quan \emph{et al.} \cite{Quan2021InOneGo} jointly removed raindrops and rain streaks with a cascaded network architecture named the CCN network. Recently, motivated by the recent progress achieved with state-of-the-art conditional generative models, Ozan \emph{et al.} \cite{ozdenizci2023Restoring} present a patch-based image restoration algorithm based on denoising diffusion probabilistic models, which achieves state-of-the-art performances on multiweather image restoration including raindrop removal. Zini \emph{et al.} \cite{Laplacian_2022_PR} design and implement a new encoder-decoder neural network first to get the Laplacian pyramid of a rain-free version of the input, making it possible to handle the variety of appearances of rain droplets. Yang \emph{et al.} \cite{Xulei2022Iter} designed an iterative feedback neural network to refine low-level representations with high-level information and gradually remove raindrops in a supervised manner and contrastive regularization. 

\subsection{Unsupervised learning raindrop removal}
Supervised learning is based on accurate and labor-intensive annotations. Weakly-supervised \cite{Nguyen2019Action, Luo2021weakly}, semi-supervised \cite{Xulei2020semi}, or unsupervised \cite{Zhu2017Unpaired, huijiao2020ffe, hj2023ffe, Xulei2019card} learning could ease this problem. Nguyen \emph{et al.} \cite{Nguyen2019Action} learn a model of attention without explicit temporal supervision by using an attention model to learn a richer notion of actions and their temporal extents. Little work focuses on unsupervised learning methods compared to supervised learning methods. To overcome the restriction of paired data, an unsupervised learning model was proposed by Luo \emph{et al.} \cite{Luo2021weakly}. They used a multitask learning manner to train a raindrop detector that could highlight raindrop regions. Then, an attention-based generative network obtained a non-raindrop image retaining details. Xulei \emph{et al.} \cite{Xulei2020semi} put forward a GAN-based semi-supervised approach, which utilizes unlabelled data to enhance the learning of data distribution and attains a superior performance in terms of AUROC under the condition of limited data. Additionally, several unsupervised methods are investigated, such as the employment of 1D CNN Autoencoder in conjunction with one-class SVM, which exhibits favorable performance in the absence of data labeling. These methods offer a viable solution to address the issue of insufficient paired data in the raindrop removal task. CycleGAN \cite{Zhu2017Unpaired} and FFE-CycleGAN \cite{huijiao2020ffe, hj2023ffe} presented a method to translate an image from a source domain to a target domain. This could be a promising method for the raindrop removal task, where images with raindrops are seen as the source domain, and those without raindrops are regarded as the target domain. 

\subsection{Layer Separation and feedback network}
Apart from the above-mentioned methods, we also took inspiration from layer separation techniques and feedback networks. Researchers use layer separation in vision tasks such as image dehazing \cite{Berman2016Non}, transparency separation \cite{Dekel2017On}, and image/video segmentation \cite{Faktor2013Co}. Double-DIP \cite{Gandelsman2019DoubleDIP} is a unified framework for the unsupervised layer decomposition of a single image. It has shown its power in many vision tasks, including segmentation, dehazing, and transparency separation. In our work, we referred to the idea that an image can be viewed as a mixture of "simple" layers.

The feedback mechanism, which was first applied to network architectures in \cite{stollenga2014deep,zamir2017feedback} and later achieved promising results for image super-resolution in \cite{li2019feedback}, was also deployed in our proposed model to refine low-level representations with high-level information gradually. In \cite{qian2018attentive}, an attention-recurrent network was designed in a feedback manner for raindrop removal. In this paper, we are the first to deploy a feedback mechanism through iterative neural networks for unsupervised single-image raindrop removal.

\begin{figure}[tb]
\centering 
\includegraphics[width=0.95\linewidth]{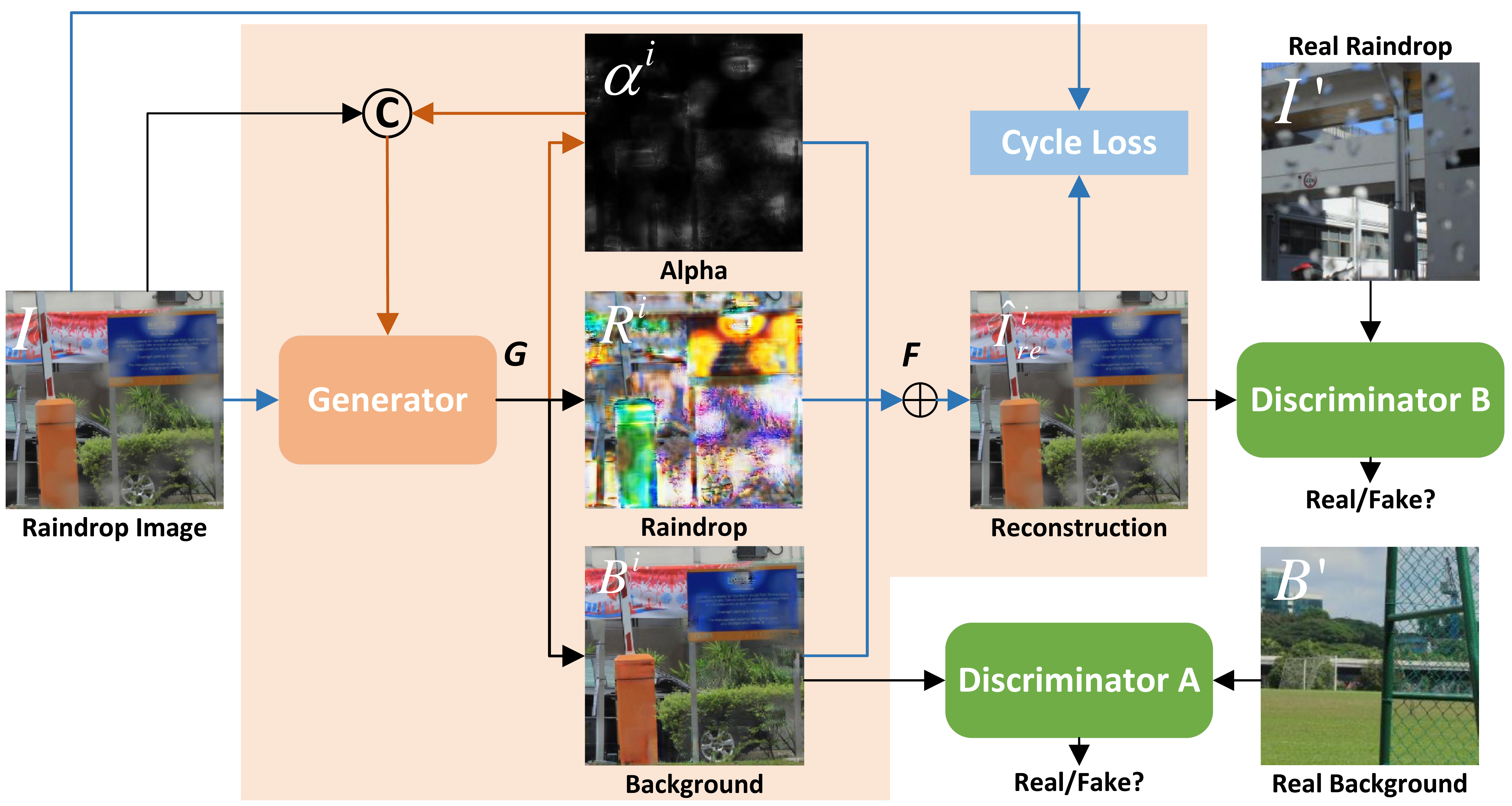}
\caption{Overall flowchart of our proposed method.} 
\label{fig31}
\end{figure}

\section{Methodology}
\label{sec:pagestyle}
Given a single rainy image, we seek to remove the raindrops and retrieve the clean background image. As shown in the overall flowchart Fig.~\ref{fig31}, our proposed method mainly consists of a cycle structure (shown in blue line) and an iterative neural network (shown in orange line) with an image decomposition mechanism. According to the image decomposition mechanism, a rainy image can be decomposed into a transparency mask, a raindrop image, and a clean background. The transparency mask will be input into the iterative training schedule combined with the input rainy image, gradually improving the decomposition accuracy. Then, we recompose the accurate transparency mask, raindrop image, and clean background to the reconstructed rainy image, which should be consistent with the input rainy image.
Meanwhile, we introduce two discriminators individually to determine whether the split clean background and the reconstructed rainy image conform to the true distribution. The clean split background is what we need in the end. The whole process does not require pairwise data to be supervised.



\begin{figure}[tbp]
\centering 
\includegraphics[width=.95\linewidth]{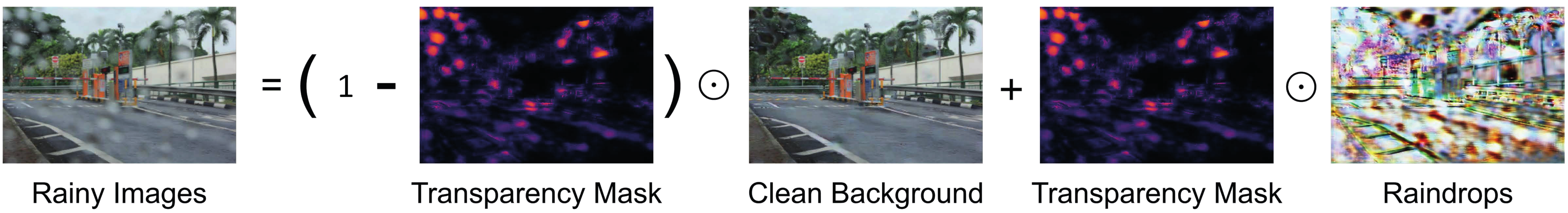}
\caption{Diagram of Image Decomposition. A heat map represents the transparency mask. The closer the value of a pixel is to 1, the closer its color is to red.} 
\label{fig2}
\end{figure}

\subsection{Model of Image Decomposition}

The task of single-image raindrop removal can be reinterpreted as decomposing the original rainy image into a background image without raindrops and an image with only raindrop effects. Such an image decomposition task model can be formulated as follows: 
\begin{equation}\label{eq1}
\begin{aligned}
    I = (1 - \alpha) \odot B + \alpha \odot R   
\end{aligned}
\end{equation}
where $I \in \mathbb{R}^{C\times H  \times W}$ denotes the image with raindrops. $B \in \mathbb{R}^{C\times H  \times W}$ represents the clean background image without the effect of raindrops. $R \in \mathbb{R}^{C\times H  \times W}$ denotes the image with only raindrop effects, composed of the raindrops themselves and their resulting distortion and blurring of the background image. $\alpha \in [0,1] ^{1\times H  \times W}$ is the transparency mask indicating the locations of the raindrops in the image. $C$, $H$, and $W$ represent the input image's number of channels, height, and width, respectively. Operator $\odot$ stands for element-wise multiplication. Fig.  \ref{fig2} illustrates an example of the above image decomposition. 
Each element in $\alpha$ represents the percentage of the original image $I$ affected by the background image $B$ and the raindrop image $R$ at the corresponding pixel. 
When the value of its element approaches 1, the corresponding pixel contains more raindrop information and less clean background image information. 

We can successfully decompose the original rainy image into the clean background image $B$ and the raindrop image $R$ with the transparency mask $\alpha$ based on such a linear model. The clean background image $B$ is the target result of single-image raindrop removal. Meanwhile, we can also reassemble the decomposed images into the original rainy images. 

\subsection{Iterative Neural Network}
\label{3.2}
\begin{figure}[tb]
\centering 
\includegraphics[width=0.95\linewidth]{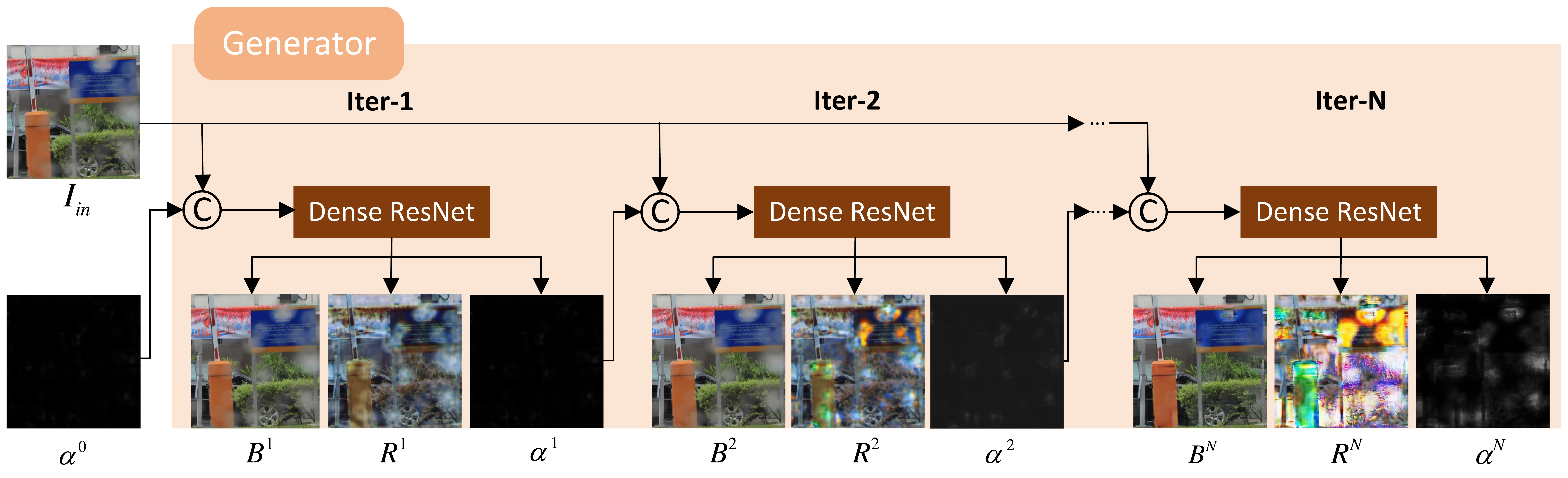}
\caption{Detailed flowchart of the iterative network as the generator in Fig. \ref{fig31}.} 
\label{fig32}
\end{figure}

Our method's proposed iterative neural network, which acted as the generator shown in Fig. \ref{fig32}, comprises $N$ iterations with the same backbone network. Feedback systems have two essential characteristics: (1) they are iterative, and (2) they reroute output back into the system. When constructing our proposed method, we 
followed a similar principle: combining high-level information output (transparent mask) from the previous iteration and the original input (rainy images) can better guide the future decomposition. This strategy enables the network to gradually learn a complex reconstruction model with only a few parameters, which enables our method to output refined images step by step.

\begin{figure*}[tb]
\centering 
\includegraphics[width=.95\linewidth]{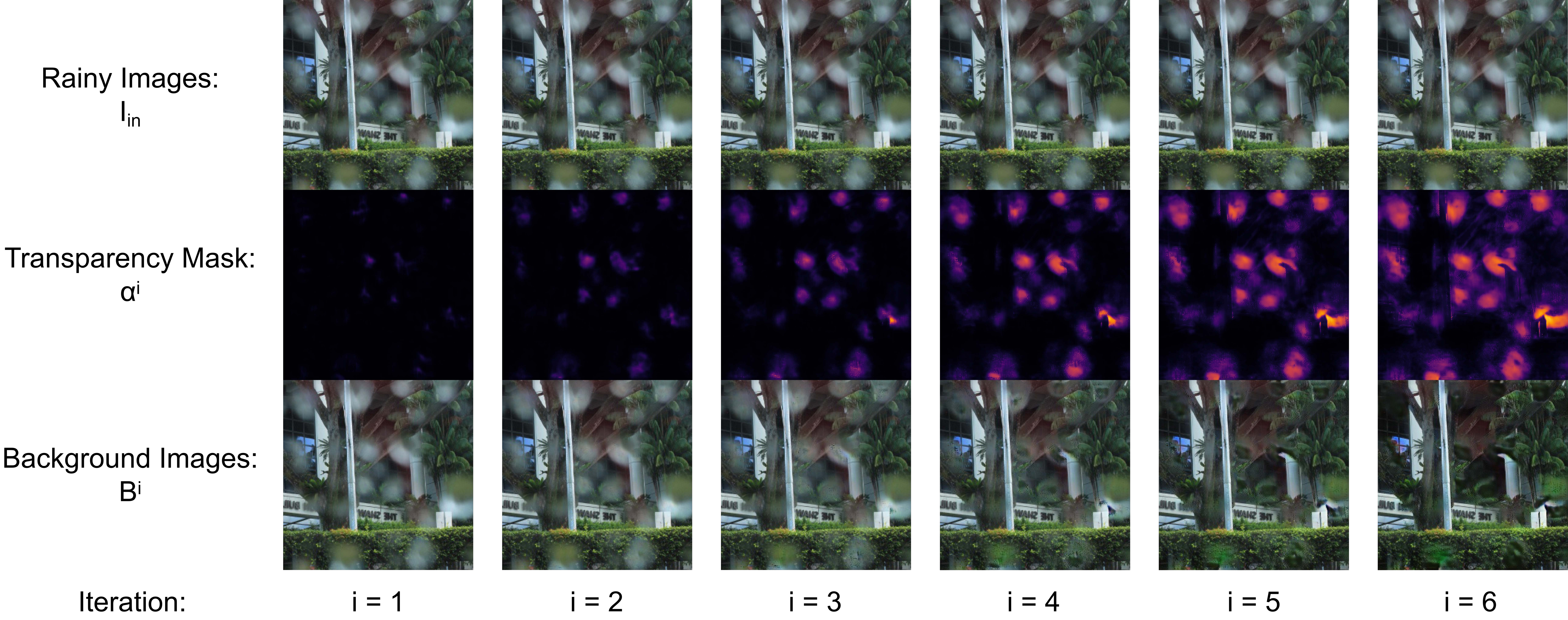}
\caption{Samples of rainy images, transparency mask, and clean background images at different iterations in our iterative neural network.   
} 
\label{fig5}
\end{figure*}

In each iteration of our proposed method, the loss function is computed independently and superimposed until the last iteration. The details of the components of our loss function and how each component operates are described in the next subsection. The sub-networks $G$ in each iteration have the same structure and share parameters with each other. Furthermore, 
the output of the current sub-network is part of the next sub-network's input.

We write the total number of iterations as $N$ and the input rainy image as $I$. Besides, we write the output clean background image as $B^i$, the transparency mask as $\alpha^{i}$, the output raindrop image as $R^i$ and the reconstructed image as $\hat{I}_{re}$ from iteration $i (i=1,2,..., N)$. Each iteration's sub-network $G$ is based on different inputs and produces different outputs. During training, firstly, we need to provide a rainy image $I$ as the input of the iterative neural network. To exploit the high-level information of the previous iteration for the next layer, we introduce direct connections from the previous iteration's output to the next iteration. Through these connections, we concatenate the transparency mask $\alpha^{i-1}$ obtained from the previous iteration $i-1$ with the original input raindrop image $I$ as the input of the next iteration $i$. Moreover, based on the image decomposition model, the sub-network $G$ output comprises the transparency mask $\alpha$, the clean background $B$, and the raindrop layer $R$. Therefore at iteration $i$, the input $G_{in}^{i}$ and output $G_{out}^{i}$ of the sub-network $G$ are formulated as: 
\begin{equation}
\begin{aligned}
    &G_{in}^{i} =  I | \alpha^{i-1} \\
    &G^{i}_{out} = B^{i}, R^{i}, \alpha^{i}
\end{aligned}
\end{equation}
where operator $|$ stands for the horizontal concatenation of two matrices. $\alpha^{0}$ is initialized as an all-zero matrix. Fig. \ref{fig5} shows the samples of rainy images, transparency masks, and clean background images at different iterations. As the number of iterations increases, the transparency mask can better detect the position and shape of raindrops, and the effect of removing raindrops is also improved.

In addition, we can obtain the reconstructed rainy images $\hat{I}_{re}^{i}$ at iteration $i$ by recombining the output $B^i$ and $R^i$ as follows:
\begin{equation}\label{eq3}
\begin{aligned}
    (1 - \alpha^{i}) \odot B^{i} + \alpha^{i} \odot R^{i} &= \hat{I}_{re}^{i}. \\
\end{aligned}
\end{equation}
This composing operation is called $F$, which is the inverse operation of Eq. \ref{eq1}.

\subsection{Cycle Structure} 
\label{3.3}


We design a GAN network with a recurrent structure for weakly -supervised single-image raindrop removal. As shown in the overall flowchart Fig. \ref{fig31}, our proposed method includes two discriminators and only one generator. 
The only generator $G$ is used for decomposing rainy images, and its specific structure is explained in detail in the Subsection \ref{3.2}. When we reconstruct the decomposed images into rainy images, our method 
performs the reconstruction by superposition of transparency masks through the function $F$. 

Meanwhile, we care more about decomposing the rainy image better for raindrop removal. Thus, we only use the decomposed image for reconstruction to constrain the decomposed domain. In this case, there is merely a single cycle in the model structure, as shown in Fig. \ref{fig31}: decompose from the input rainy image to generate two image layers and a mask and then combine the decomposition results to reconstruct the rainy image. The reconstructed rainy image should be consistent with the input image.

Under such a framework, we employ two discriminators for the adversarial training of the generator. The discriminator $D_A$ is used to discriminate the authenticity of the clean background images $B^i$ decomposed by the input original rainy images, while the other discriminator $D_B$ is responsible for distinguishing the restored rainy images $\hat{I}_{re}^{i}$. They adopt the same structure in PatchGAN \cite{demir2018patch}. The discriminator in PatchGAN is designed as a fully convolutional structure, which uses convolution to map the output to a matrix to evaluate the generated image.

As shown in Fig. \ref{fig31}, we write the distribution of another true clean background as $B^{'} \sim P_{data} (B)$ and the distribution of another true rainy image as $I^{'} \sim P_{data} (I)$, to distinguish them from the input rainy image $I$ and the ground truth $B$ of $B^{i}$. Therefore, the following adversarial losses can be expressed:
\begin{equation}
\begin{aligned}
    L_{GAN} (D_A, B^{'}, B^{i}) &= E_{B^{'} \sim P_{data} (B)} [log D_A (B^{'})] \\
    &+ E[log (1-D_A (B^{i}))]\\
    L_{GAN} (G, F, D_B, I, I^{'} ) &= E_{I^{'} \sim P_{data} (I)} [log D_B (I^{'})] \\
    &+ E_{I \sim P_{data} }(I)[log (1-D_B (F(G(I)))]
\end{aligned}
\end{equation}

However, only with the constraint of adversarial loss can the network not guarantee that the image's content before and after transformation is consistent. Therefore, we need the following cycle loss to ensure the consistency of the content, that is, to narrow the discrepancy between the reconstructed rainy image from the decomposed image and the original rainy image in our method:
\begin{equation}
\begin{aligned}
    L_{cyc} (G,F ) &= E_{I \sim P_{data}(I)} [|| F(G(I))-I ||_1] \\
\end{aligned}
\end{equation}

Moreover, the generator $G$ is used to generate clean background images $B^i$, then sending $B$ to $G$ should still generate $B$.
Only in this way can we prove that $G$ can generate clean background images. Hence, $B^{i}$ and $B$ should be as close as possible. Identity loss is defined as: 
\begin{equation}
\begin{aligned}
    L_{identity} (B^{i}, B) &= E_{B \sim P_{data}(B)} [|| B^{i}-B ||_1] \\
\end{aligned}
\end{equation}


\subsection{Loss Function}

During a training session, each iteration in the iterative neural network is tied with a loss function and produces one decomposed image group. Moreover, decomposed images are reconstructed into a rainy image in each iteration through Eq. \ref{eq3}. The total adversarial loss function comprises all the decomposed background images’ and reconstructed rainy images' loss functions output by N iterations. Hence, $L_{GAN}$ can be formulated as:
\begin{equation}
    L_{GAN}(G, F, D_A, D_B) = \sum_{i}^{N} K_i \frac{L_{GAN} (D_A, B^{'}, B^{i}) + L_{GAN} (G, F, D_B, I, I^{'} ) }{2}
\end{equation}
since the background image decomposed by the iterative neural network will get closer and closer to the desirable clean background image as the iteration number increases, we set $K_i=i-1$ to assign more weights for later iterations.

Aside from the adversarial loss, cycle consistency loss, and identity Loss, we formulate an additional loss on $\alpha$ for learning an accurate transparency mask that segments out the raindrops. In the rainy images, it's observed that raindrops only take up a small portion, and the rest is the background. The constraints based on this fact allow us to define the following loss function:

\begin{equation}
\begin{aligned}
    L_{sparsity}(\alpha) &=  L_1(\alpha,0) \\
\end{aligned}
\end{equation}
this means most of the values on $\alpha$ should be close to 0 since a major portion of the rainy image is the background. This could prevent the loss of smoothness in $\alpha$ since the occurrence number of 1 is now constrained.

Combing cycle consistency loss $L_{cyc}$, identity loss $L_{identity}$, GAN loss $L_{GAN}$, and Sparsity Loss $L_{sparsity}$, the final loss function of our proposed model is formulated by
\begin{equation}
    L_{total} = \beta_{1} L_{GAN}(G, F, D_A, D_B) + \beta_{2} L_{cyc}(G, F) + \beta_{3} L_{identity}(B^{i}, B) +\beta_{4} L_{sparsity}(\alpha)
\end{equation}
where the weights $\beta_{1\sim4}$ are introduced as hyper-parameters and determined by experiments.

\section{Experiments}
\label{sec:exp}
In this section, we evaluate our proposed network quantitatively and qualitatively. Not only do we perform the ablation study, but also we compare our network with the state-of-the-art unsupervised methods on two benchmark datasets. 
There are four unsupervised methods used in our comparisons: an attention-based translation model called UA\_GAN \cite{alami2018unsupervised}, an image-to-image translation model using unpaired data named AGGAN \cite{tang2019attention}, a general unpaired image translation method called CycleGAN \cite{Zhu2017Unpaired}, and an unsupervised model \cite{Luo2021weakly}. Note that UA\_GAN, AGGAN, and CycleGAN are general image-to-image translation methods and are not dedicated to raindrop removal. The unsupervised model \cite{Luo2021weakly} is the only one we can find in the literature specifically designed for raindrop removal. 

\subsection{Experiments Setup}
\textbf{Datasets and Metrics} Although we train our network with unpaired images, evaluating the quality of generated clear background images needs the corresponding ground truth images. Therefore, we need a paired dataset to test our model. To fully verify the performance of our proposed model, we do experiments on two different datasets: the NUS raindrop dataset released by Qian \emph{et al.} \cite{qian2018attentive} and the RainDS dataset \cite{Quan2021InOneGo}. These datasets include numerous image pairs in various lighting conditions and different scenes. Following previous deraining methods, the quantitative evaluation uses two metrics: the Peak Signal Noise Ratio (PSNR) and the Structural Similarity Index (SSIM) calculated on the processed image. 

\textbf{Implementation Details}  
We adopt the PyTorch library to implement our model and conduct raindrop removal experiments on the NVIDIA platform with GeForce GTX TITAN X cards. We borrow the network structure of CycleGAN to train our model. We train the model on the training set and evaluate the performance on the evaluation set. During training, a patch-based strategy is used to crop patches of 256x256 from each image randomly. The quantitative evaluation uses two metrics: PSNR and SSIM, calculated on separated backgrounds and given GT images. Each model is trained for 400 epochs on 2 GPUs with a total batch size of 6. The base learning rate is 0.001. All models are trained from scratch and are optimized using stochastic gradient descent with momentum at 0.9 and weight decay of 1e-5.  The hyper-parameters $\beta_{1\sim4}$ are $2 \times 1.5^{i-1}$, 10, 5, and 1 respectively. The $i$ in $2 \times 1.5^{i-1}$ is the iteration number of $IterNN$ in Fig. \ref{fig5}, which is from 1 to 6.

\subsection{Compared with the State-of-the-Arts on NUS Raindrop Dataset}

\begin{figure*}[!ht]
\centering
\includegraphics[width=0.99\linewidth]{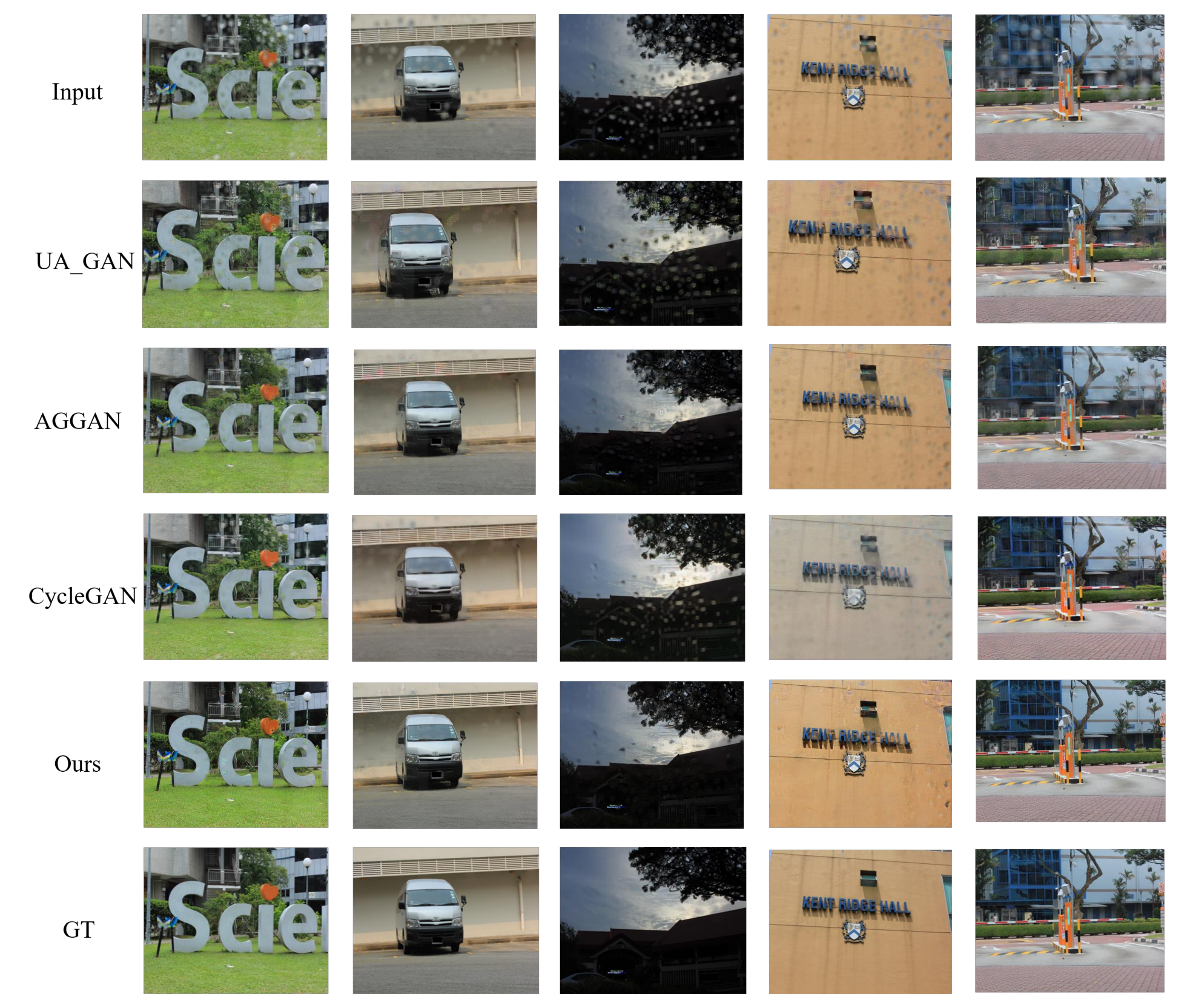}
\caption{Results on NUS Raindrop dataset. From top to bottom: the input raindrop image, UA\_GAN, AGGAN, CycleGAN, our model, and the ground truth. }
\vspace{-0.2in}
\label{figNUS}
\end{figure*}

NUS raindrop dataset \cite{qian2018attentive} has been widely used in recent years. The images with raindrops in the NUS raindrop dataset are created like Eigen \emph{et al.} \cite{Eigen2013dirt}: taking photos before/after spraying water on the glass. The NUS raindrop dataset contains 1119 corrupted/clean image pairs. Following the same strategy as \cite{qian2018attentive}, we use 861 image pairs for training. Image patches of size $256 \times 256$ are randomly cropped from the training images as the input/truth data. In the remaining dataset images, 58 well-aligned image pairs are picked out for test\_a and 249 image pairs for test\_b.

In this section, we compare our method with the four mentioned unsupervised methods: CycleGAN \cite{Zhu2017Unpaired}, UA\_GAN \cite{alami2018unsupervised}, AGGAN \cite{tang2019attention}, weakly supervised model \cite{Luo2021weakly}. Following the dataset split settings in the NUS raindrop dataset \cite{qian2018attentive}, we retrain the CycleGAN, UA\_GAN, and AGGAN models. Since the unsupervised model hasn't released the source codes, we directly quote the quantitative results from its paper \cite{Luo2021weakly}.

\textbf{Qualitative Evaluation}
In Fig. \ref{figNUS}, we present the retrieved clean images by our model, CycleGAN \cite{Zhu2017Unpaired}, UA\_GAN \cite{alami2018unsupervised}, and AGGAN \cite{tang2019attention}. The background images generated by our model are less artificial. Compared to CycleGAN \cite{Zhu2017Unpaired}, the color distribution of our results is closer to the real image. We can observe that UA\_GAN and CycleGAN cannot successfully remove the dense rain. However, the raindrops in our results have been removed, whatever their shape, size, and density. And the region blocked by raindrops is retrieved with more real information.

\begin{table}[!htb]
\renewcommand{\arraystretch}{1.}
\caption{Quantitative evaluation results on NUS Raindrop Dataset  \cite{qian2018attentive}. }
\begin{center}
\setlength{\tabcolsep}{3.mm}
\begin{tabular}{l|c|c|c|c}
\hline
\multirow{2}{*}{Methods} & \multicolumn{2}{c}{test\_a}& \multicolumn{2}{|c}{test\_b } \\
\cline{2-5}   
 & PSNR & SSIM & PSNR & SSIM \\    
\hline
Baseline & 23.8099 & 0.8363 & 21.1634 & 0.7489 \\
UA\_GAN \cite{alami2018unsupervised}  & 23.9678 & 0.7698  & 21.6782  & 0.6896  \\
CycleGAN \cite{Zhu2017Unpaired}  & 23.7620 & 0.8334 & 22.0835 & 0.7933 \\
AGGAN \cite{tang2019attention}  & 25.0985 & 0.8218  & 22.7200  &  0.7382 \\
Weakly \cite{Luo2021weakly} & 25.4624 & \textbf{0.8763} & 23.2445 & 0.8064 \\
Ours & \textbf{27.0562} & 0.8738 & \textbf{24.7124} &\textbf{0.8281} \\  
\hline
\end{tabular}
\end{center}
\vspace{-0.2in}
\label{QuanNUS}
\end{table}

\textbf{Quantitative Evaluation} Table \ref{QuanNUS} shows the results of our model and other state-of-the-art methods. We can see that our proposed method outperforms all the other methods, especially on PSNR. To our knowledge, the unsupervised model \cite{Luo2021weakly} is the most recent model designed for raindrop removal training on unpaired datasets. Our method surpasses the unsupervised model by 1.5938 dB of PSNR on test\_a and 1.4679 dB on test\_b. We obtained comparable SSIM values with the unsupervised model on test\_a. While on the harder dataset test\_b, our model outperforms the unsupervised model by 0.0217 on SSIM. 

\subsection{Compared with the State-of-the-Arts on RainDS dataset}
\begin{figure}[t]
\centering
\includegraphics[width=0.99\linewidth]{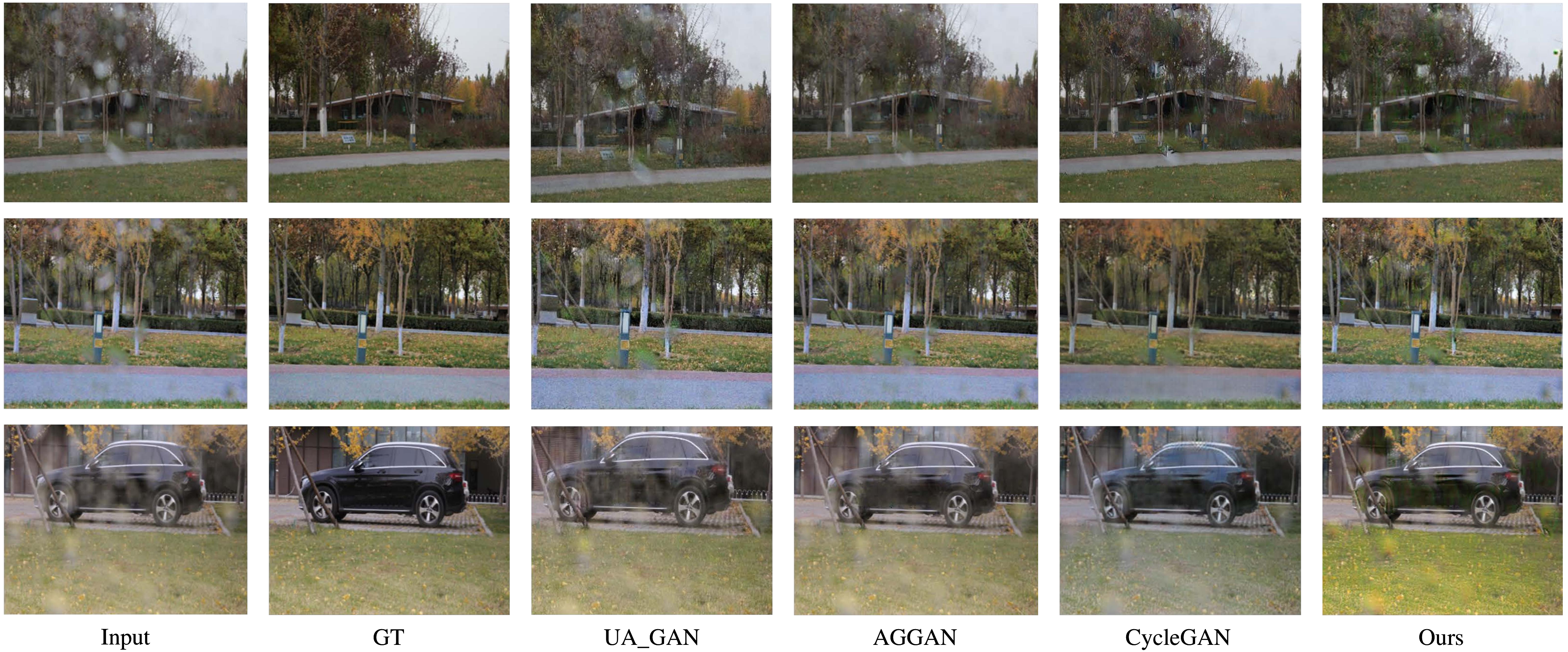}
\caption{Results on RainDS dataset. From left to right: the input raindrop image, the ground truth, UA\_GAN, AGGAN, CycleGAN, and ours. }
\vspace{-0.2in}
\label{figRainDS}
\end{figure}

To further evaluate our model on various raindrop data and test the robustness, we compare our model with the state-of-the-art methods on the RainDS dataset \cite{Quan2021InOneGo}. The RainDS dataset includes numerous image pairs in various lighting conditions and scenes. RainDS consists of real-world rain images and a synthetic subset generated in autonomous driving scenes. It is collected by Quan et al. \cite{Quan2021InOneGo}. Each image pair contains four images: a rain streak image, a raindrop image, an image including both types of rain, and their rain-free counterparts. We use raindrop and rain-free images to train and test following the original split setting. The images from the real world are called RainDS\_real for short, and the synthetic images are called RainDS\_syn for short. There is a big difference between the generated raindrop data in RainDS\_syn and the raindrop image in the real scene because raindrops and backgrounds have different focal lengths in real images. Therefore, we only do experiments on RainDS\_real to extensively evaluate the performance of our proposed network. 
RainDS\_real has 248 paired samples, i.e., clean background and raindrop images, where 150 pairs are selected for training and 98 pairs for testing. 
In the following, we use RainDS to substitute RainDS\_real for convenience.

\textbf{Qualitative Evaluation}

In Fig. \ref{figRainDS}, it is obvious that the background images generated by our model are less artificial. Compared to CycleGAN \cite{Zhu2017Unpaired}, AGGAN \cite{tang2019attention}, and UA\_GAN \cite{alami2018unsupervised}, our model generates the most natural background images, achieving the similar color and texture details as the ground truth. However, CycleGAN, AGGAN, and UA\_GAN suffer from slight color distortion or texture loss and cannot completely remove the raindrops.

\textbf{Quantitative Evaluation} Table \ref{QuanRainDS}  shows the quantitative results of our model and other unsupervised methods. Our method outperforms all other methods with 21.7624dB PSNR and 0.7472 SSIM on the RainDS dataset. This shows that our proposed model is robust and can perform well on different raindrop data.

\begin{table}[tb]
\renewcommand{\arraystretch}{1.}
\caption{Quantitative evaluation results on RainDS Dataset\cite{Quan2021InOneGo}. }
\begin{center}
\setlength{\tabcolsep}{3.mm}{
\begin{tabular}{l|c|c}
\hline
Methods & PSNR & SSIM  \\   
\hline
Baseline & 18.8484 & 0.5719 \\
UA\_GAN \cite{alami2018unsupervised}  & 19.7235 & 0.6535 \\
CycleGAN \cite{Zhu2017Unpaired}  & 20.3807 & 0.6904 \\
AGGAN \cite{tang2019attention}  & 21.5399 & 0.7450 \\ 
Our model &  \textbf{21.7624} & \textbf{0.7472} \\
\hline
\end{tabular}} 
\end{center}
\vspace{-0.2in}
\label{QuanRainDS}
\end{table}

\begin{figure*}[tbp]
\centering
\includegraphics[width=1\linewidth]{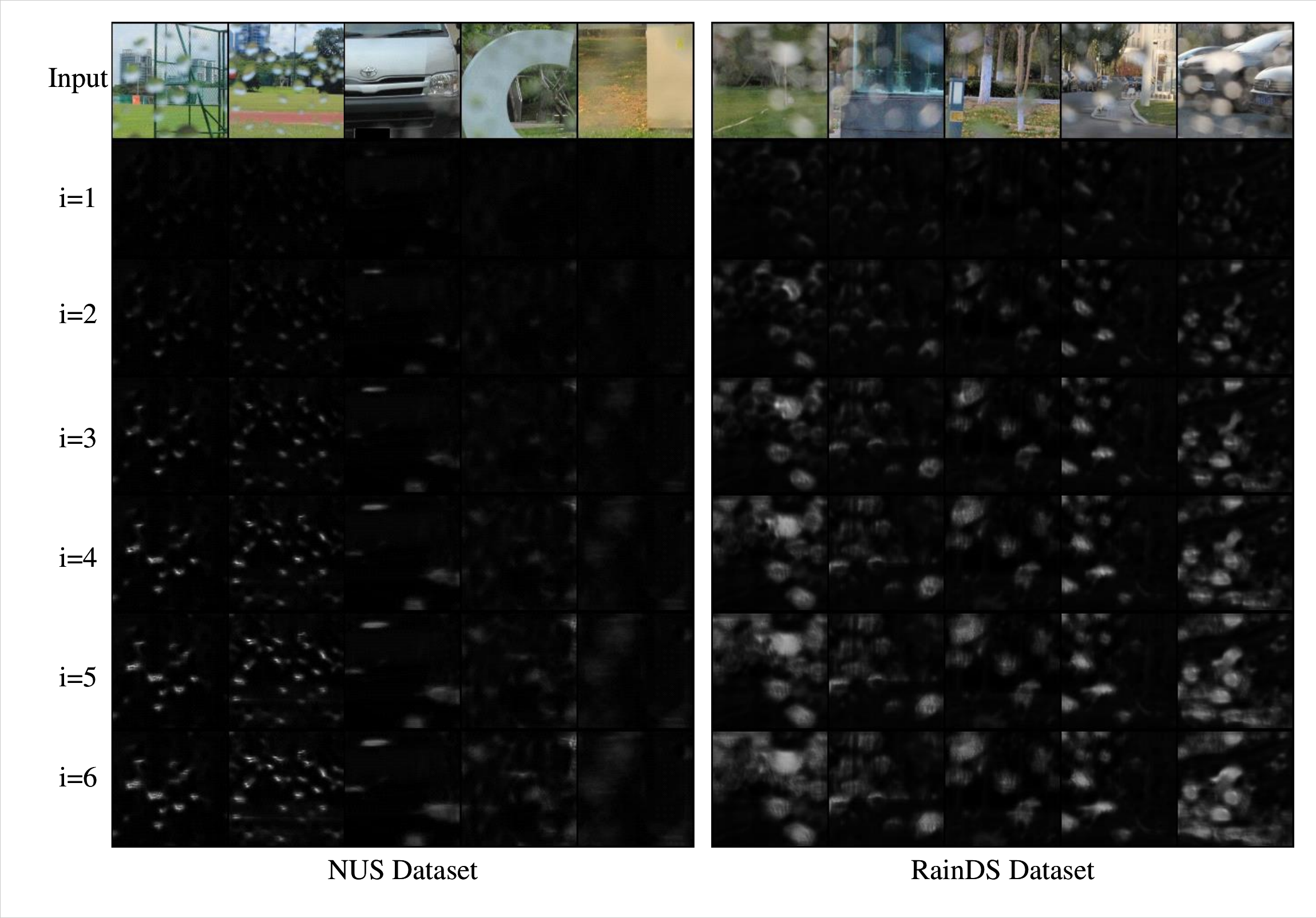}
\caption{Visualization of intermediate output from feedback mechanism on NUS Raindrop dataset and RainDS dataset. From top to bottom: the input raindrop image, the learned transparency mask from iteration $i$ from 1 to $N$. }
\vspace{-0.2in}
\label{figiter}
\end{figure*}

\subsection{Feedback Mechanism via Iterative Neural Network}

The feedback strategy is introduced with the specific aim of systematically capturing the structural intricacies of raindrops across different scales within a single input. As illustrated in Fig. \ref{figiter}, this strategic approach prominently refines low-level information with high-level details in a top-down fashion. This refinement process facilitates the gradual acquisition of raindrop knowledge, progressing from coarse to fine granularity. Regardless of whether the raindrop is large or small, sparse or dense, and the variation in their shapes, the results show that the model can learn the location, shape, and thickness of the raindrop. The pivotal role of the feedback mechanism in this progressive learning process ensures a comprehensive and nuanced understanding of raindrops.

\subsection{Ablation Study}
In this section, we explored the performance of different modules in the proposed method by conducting ablation experiments on the NUS raindrop dataset \cite{qian2018attentive}. As mentioned in the methodology section, we proposed an iterative neural network, the cycle consistency loss $L_{cyc}$, the identity loss $L_{identity}$, and the sparsity loss $L_{sparsity}$. The details of the variants are listed below.
\begin{itemize}
    \item $w/o $  $L_{cyc}$ : Removing the cycle consistency
loss ($L_{cyc}$) from the proposed network during training procedure;
    \item $w/o $  $L_{identity}$ : Removing the identity loss ($L_{identity}$) from the proposed network during training procedure;
    \item $w/o $  $L_{sparsity}$ : Removing the sparsity loss $L_{sparsity}$ on the transparency mask $\alpha$ from the proposed network during training procedure;
    \item $w/o $  $IterNN$ : Removing the iterative neural network ($IterNN$ for brief) from the proposed network.
\end{itemize}
We explore all these variants both quantitatively and qualitatively.

\begin{figure*}[!ht]
\centering
\includegraphics[width=0.99\linewidth]{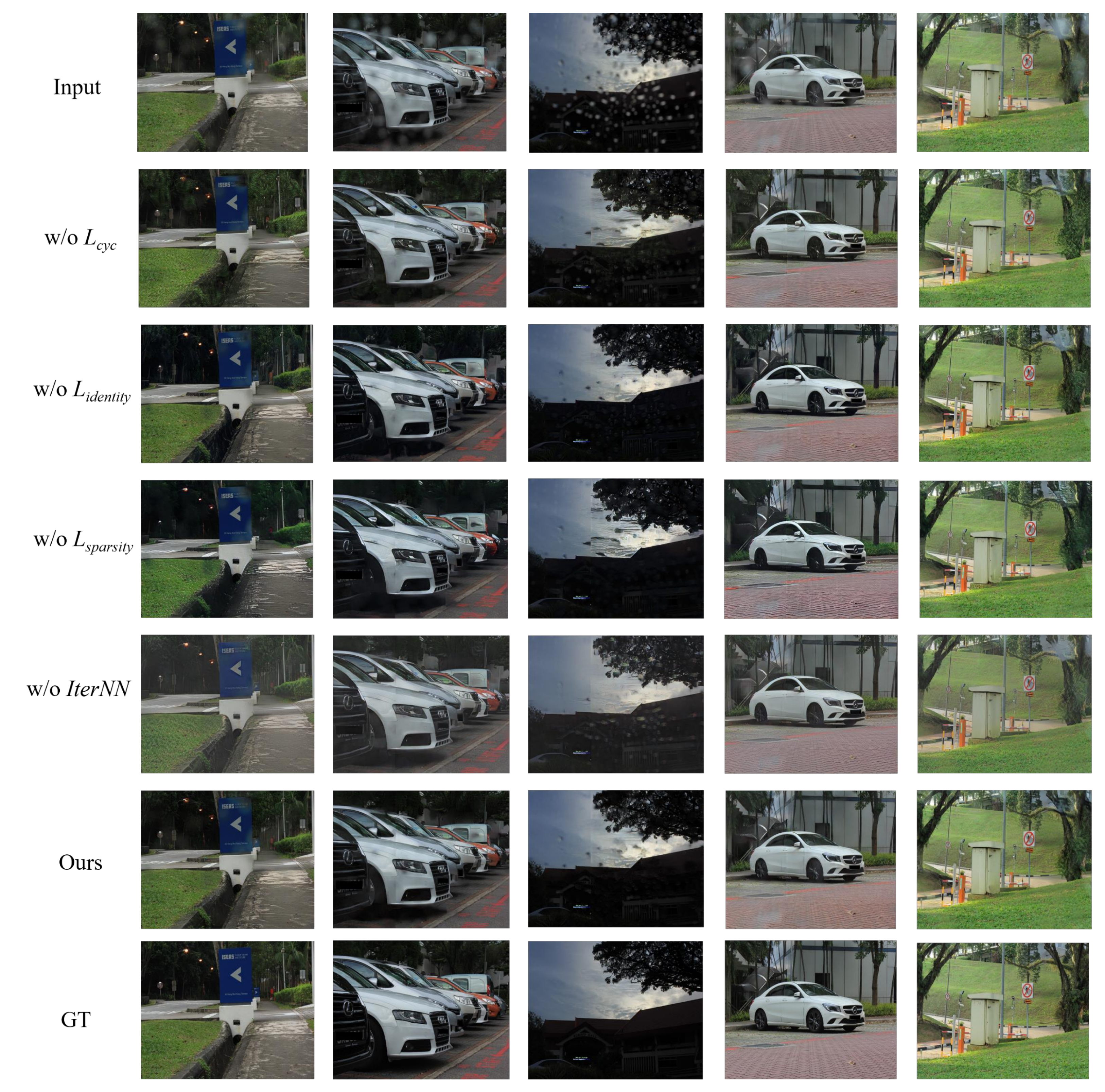}
\caption{Component analysis. Results of the proposed model and its variants.}
\label{figNUSA}
\end{figure*}

\begin{table}[!tb]
\renewcommand{\arraystretch}{1.}
\caption{Quantitative evaluation results of ablation study on Raindrop Dataset \cite{qian2018attentive}. }
\begin{center}
\setlength{\tabcolsep}{3.mm}
\begin{tabular}{l|c|c|c|c}
\hline
\multirow{2}{*}{Methods}& \multicolumn{2}{c}{test\_a}& \multicolumn{2}{|c}{test\_b } \\
\cline{2-5}   
 & PSNR & SSIM & PSNR & SSIM \\    
\hline
$w/o $  $L_{cyc}$ & 24.7130 & 0.7825 & 21.8595 & 0.7021 \\
$w/o $  $L_{identity}$ & 25.8741 & 0.8348 & 22.6429 & 0.7448 \\
$w/o $  $L_{sparsity}$ & 25.1320 & 0.8317 & 22.5002 & 0.7432\\
$w/o $  $IterNN$ & 25.2506 & 0.8411 & 22.0639 & 0.7518\\ 
Ours (Full model) & \textbf{27.0562} & \textbf{0.8738} & \textbf{24.7124} &\textbf{0.8281}\\  
\hline
\end{tabular}
\end{center}
\vspace{-0.2in}
\label{abTable}
\end{table}

\textbf{Quantitative Evaluation.} Table \ref{abTable} shows the PSNR and SSIM results of different variants of the proposed network on NUS raindrop test\_a and test\_b \cite{qian2018attentive}. Comparing all variants with our complete proposed model, it can be seen that all of them can bring more or less improvement. Among them, the most efficient loss is $L_{cyc}$, which improves PSNR by 2.2475 dB and SSIM by 0.1260 on test\_b. And the contribution of $L_{sparsity}$ takes the second place. It improves PSNR by 1.9242 dB on test\_a and SSIM by 0.0849 on SSIM on test\_b compared to the full architecture. $L_{identity}$ and $IterNN$ also play important roles in improving PSNR and SSIM on test\_b. Test\_b contains more images and is harder than test\_a. However, the improvement on test\_b is more obvious, indicating the effectiveness of our proposed different modules.

\textbf{Qualitative Evaluation}
To evaluate the effectiveness of different variants more intuitively, the qualitative results on NUS test\_a are shown in Fig. \ref{figNUSA}. The variants remove most of the raindrops at different degrees. The results of $w/o$ $L_{sparsity}$ look very rough, and the texture of the images is full of artifacts. The reason is that the constraint of $L_{sparsity}$ urges the $\alpha$ to learn an accurate transparency mask. Therefore, without the constraint of $L_{sparsity}$, the qualification of the generated clean images decays the most. $L_{cyc}$ and $L_{identity}$ are both trying to retrieve the blocked information. Hence, we can also obverse that the results of $w/o$ $L_{cyc}$ and $w/o$ $L_{identity}$ show an oil painting-like texture in the restored part, losing a lot of details. Comparing the results of $w/o$  $IterNN$ and our full model, it is easy to observe that the $IterNN$ module effectively emits artificial details and learns more background details. Therefore, the proposed modules are very efficient in improving the visual performance of raindrop removal.

\section{Conclusion}
We propose an unsupervised network for low-level vision tasks - raindrop removal from a single image. The proposed model performs layer separation based on cycle and feedback network structure. Firstly, in a cycleway, the specifically designed loss functions, i.e., GAN Loss, cycle consistency loss, identity loss, and sparsity loss, are used to guide the separation of rainy images into clean and rainy layers. Secondly, in an iterative way, the feedback mechanism is deployed to refine low-level representation with high-level information gradually. Experimental results on two real raindrop benchmark datasets show the effectiveness of the proposed method. We hope this work can motivate further research on designing efficient network structures for unsupervised raindrop removal.

\printcredits

\bibliographystyle{cas-model2-names}

\bibliography{cas-refs}

\bio{}
\endbio


\end{document}